\documentclass[letterpaper, 10 pt, conference]{ieeeconf}  

\IEEEoverridecommandlockouts 

\usepackage{amsmath,amsfonts,amssymb,mathtools} 
\usepackage{eucal,bbm,bm,nicefrac} 
\usepackage{graphicx,float} 
	\graphicspath{{./images/}}
\usepackage[table,xcdraw]{xcolor}
\usepackage{cases}
\usepackage{bm}
\usepackage{soul,color}

\newtheorem{remark}{Remark}

\newcommand{\T}{\ensuremath{\mathrm{T}}} 
\newcommand{\floor}[1]{\left\lfloor #1 \right\rfloor}

\title{\LARGE \bf SLAS: Speed and Lane Advisory System for Highway Navigation}

\author{Faizan M. Tariq$^{1}$, David Isele$^{2}$, John S. Baras$^{1}$ and Sangjae Bae$^{2}$
\thanks{$^{1}$University of Maryland, College Park, MD, USA. Email: \tt\small\{mftariq,baras\}@umd.edu.}
\thanks{$^{2}$Honda Research Institute, San Jose, CA, USA. Email: \tt\small\{disele,sbae\}@honda-ri.com.}
\thanks{Research supported by Honda Research Institute, USA.}
}

\begin{document}

\maketitle
\thispagestyle{empty}
\pagestyle{empty}
\begin{abstract}
This paper proposes a hierarchical autonomous vehicle navigation architecture, composed of a high-level speed and lane advisory system (SLAS) coupled with low-level trajectory generation and trajectory following modules. Specifically, we target a multi-lane highway driving scenario where an autonomous ego vehicle navigates in traffic. We propose a novel receding horizon mixed-integer optimization based method for SLAS with the objective to minimize travel time while accounting for passenger comfort. We further incorporate various modifications in the proposed approach to improve the overall computational efficiency and achieve real-time performance. We demonstrate the efficacy of the proposed approach in contrast to the existing methods, when applied in conjunction with state-of-the-art trajectory generation and trajectory following frameworks, in a CARLA simulation environment.
\end{abstract}

\section{Introduction}

Lane changing is considered to be one of the most risky driving behaviors since it is highly contingent upon multi-modal trajectory predictions of neighboring vehicles and requires timely decision making \cite{laneChangingRisk}. It is further influenced by a number of uncertainty factors such as road conditions, measurement accuracy, and a long tail of behavioral uncertainty of on-road agents. However, if executed efficiently, lane changing coupled with speed adjustment can yield significant improvement in minimizing overall travel time while ensuring passenger comfort \cite{bae2021risk}.

\begin{figure} [ht]
\centering
\includegraphics[trim=0 0 0 0, clip,width=.485\textwidth]{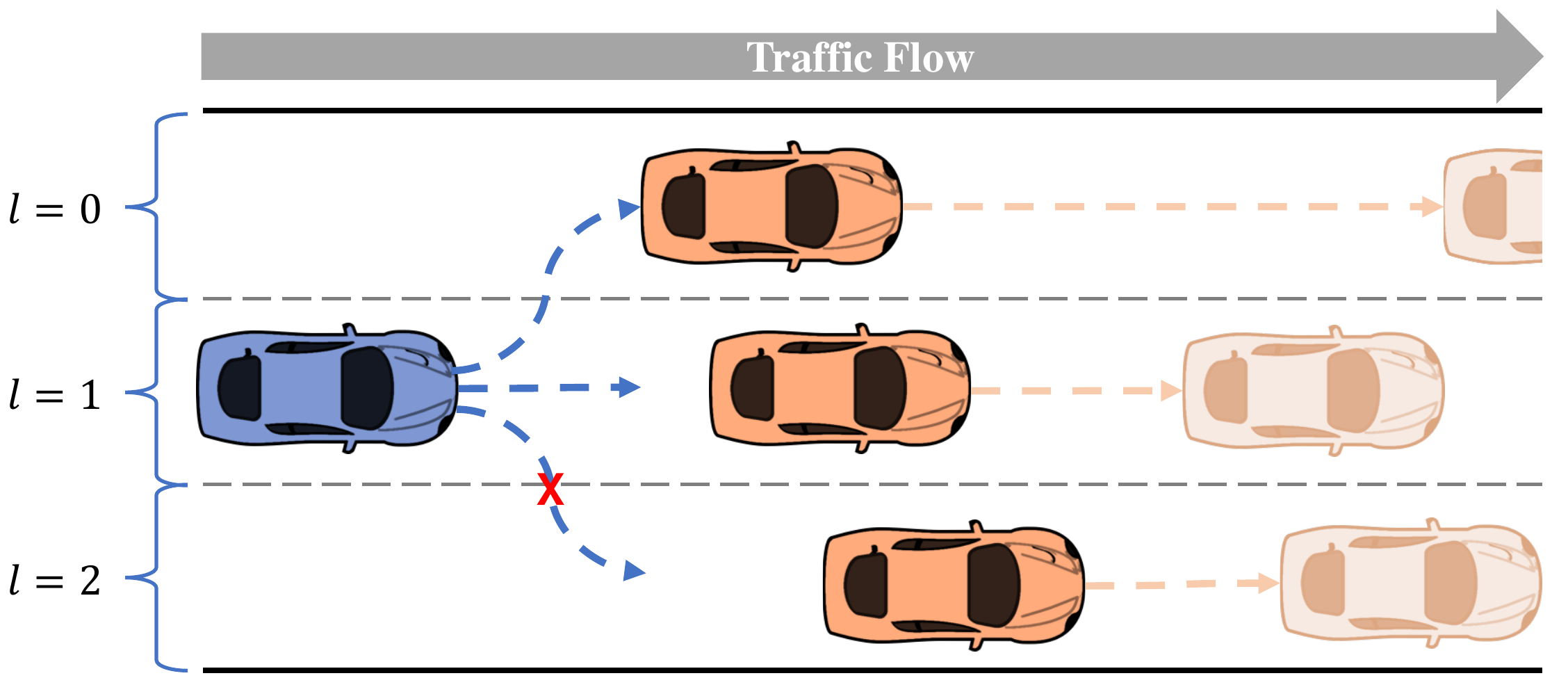}
\caption{\textbf{Motivational Example}. With a slow moving vehicle ahead, the ego vehicle (in blue) may decide to either change lane to the fast moving lane (left) to minimize travel time or adjust its speed without changing lanes to preserve safety but it would be unwise for it to switch to the slow moving lane (right) as that would not benefit travel time or safety.}
\label{fig:overview}
\vspace{-15pt}
\end{figure}

To elaborate further, consider the scenario presented in Fig. \ref{fig:overview}. Based on the predicted motion (shown in a lighter shade) of the neighboring vehicles (shown in orange), the ego vehicle (shown in blue) may decide to either change lane left in an attempt to minimize its travel time or slow down in the current lane to maintain safety. However, it would be imprudent for the ego vehicle to risk changing lane right and consequently get stuck behind a slow moving vehicle even though there is presently a greater headway. This simple scenario highlights the importance of foresight and long planning-horizon in strategic decision making for autonomous vehicles.

Existing methods like MOBIL \cite{mobil} give us the ability to change lanes but behave greedily (prioritizing immediate rewards) oftentimes, which can lead to sub-optimal performance. It was shown in \cite{bae2021risk} that the lane changing performance can be improved with an A$^\star$ inspired approach, but the formulation was limited to constant speed. Such an approach is unable to assess the benefits of speed adjustment in minimizing overall travel time. As will become apparent in Section \ref{sec:results}, it may be necessary at times to sacrifice on short-term benefits to gain long-term performance improvements. In such a scenario, an approach with speed adjustment coupled with long planning horizon has the foresight to deliver significantly better results. Moreover, the inclusion of speed adjustment in the decision making process inhibits the risk of incurring trajectory infeasibility as the environment conditions may prevent the ego vehicle from traveling at a constant reference speed and the low-level planner may be unable to handle such a discrepancy. Therefore, in this work, we propose a low complexity receding horizon optimization based approach that outputs the lane change maneuvers coupled with speed adjustments for long planning horizons ($> 15s$) while guaranteeing safety. The long horizon strategic decision making gives ego vehicle the ability to proactively anticipate and handle challenging driving situations.


\subsubsection*{Literature review}
In the literature, speed and lane changing decisions are generally considered from a motion planner's perspective \cite{motionPlanningLaneVelocity}, which allows for a simultaneous determination of target lanes and waypoints to perform the maneuver. The motion planning methods present in the literature can broadly be categorized into sampling-based, learning-based and optimization-based approaches.

In regards to the sampling-based approaches, single-query methods, in particular the different variants of RRT, are preferred over multi-query methods, like roadmap-based methods, due to the faster execution time and their ability to incorporate non-holonomic constraints \cite{rrt}. Even though these methods are able to incorporate safety guarantees by sampling feasible trajectories from a reachable safe set \cite{rrtFrazzoli}, the overall driving experience is often rather uncomfortable due to the concatenation of individual trajectories. Moreover, the asymptotic optimality guarantees availed by these methods do not help with real-world implementation in complex driving scenarios since they tend to have high sample complexity \cite{rrt}.

In terms of the learning-based methods, the preferred approach seems to be the variations of Reinforcement Learning techniques applied in a simulated environment \cite{mukadam2017tactical,dqn,drl,yang2020cm3,saxena2020driving}. These approaches, although seeming to work well in simulation, have concerns regarding real-world implementation due to the large amount of training data that they require, the exploration of unsafe behaviors during training, and a general inability to handle edge cases. They mainly utilize neural networks as function approximators which yields low computational complexity but also results in a lack of explainability and safety guarantees.

Lastly, the optimization-based approaches, especially the derivatives of optimal control methods, are abundant in the literature. In contrast to the potential-field based approaches \cite{potential} that yield decent collision avoidance performance but are unable to accommodate vehicle dynamics, the optimal control methods \cite{optimalControl}, especially the derivatives of Model Predictive Control (MPC) approach \cite{mpcDriving, mpcPlanning, scenarioMPC}, yield excellent collision avoidance performance while accommodating vehicle dynamics. However, this performance comes at a cost of high computational complexity, arising mainly from the non-convex collision avoidance and the non-linear dynamics constraints. This, in turn, restricts the planning horizon to merely a few seconds.

\subsubsection*{Contribution}

The key requirements for the algorithmic design of an autonomous vehicle include real-time operation, safety guarantees, optimality with respect to some metric(s), and accounting for the behavior variability of on-road agents. Considering these requirements, we propose an optimization-based behavioral planning framework that enables autonomous vehicle maneuvering on multi-lane highways. While having the benefits of optimization-based approaches, our method achieves a low computational complexity by employing a binary representation of the decoupled lane indicator dynamics in lieu of lateral dynamics, and utilizing algorithmic modifications to aid numerical computations. Specifically, our method provides:
\begin{itemize}
    \item optimality with respect to travel time and comfort;
    \item safety and feasibility guarantees;
    \item real-time applicability for a long planning horizon; and
    \item modularity in design, which enables the integration of external trajectory prediction modules.
\end{itemize}
The proposed method fills in the research gap by meeting all the key algorithmic requirements while simultaneously gaining the foresight to make strategic decisions that yield long-term performance benefits, as verified in Section \ref{sec:results}.

\section{Problem Formulation}
In this section, we present the algorithmic pipeline and formalize the road, observation and vehicle dynamics models that will be utilized in the subsequent sections.

\vspace{-0.8pt}

\subsection{Algorithmic Pipeline}
\label{sec:pipeline}

\begin{figure} [ht]
\centering
\includegraphics[trim=0 0.1cm 0 0.1cm, clip,width=0.9\columnwidth]{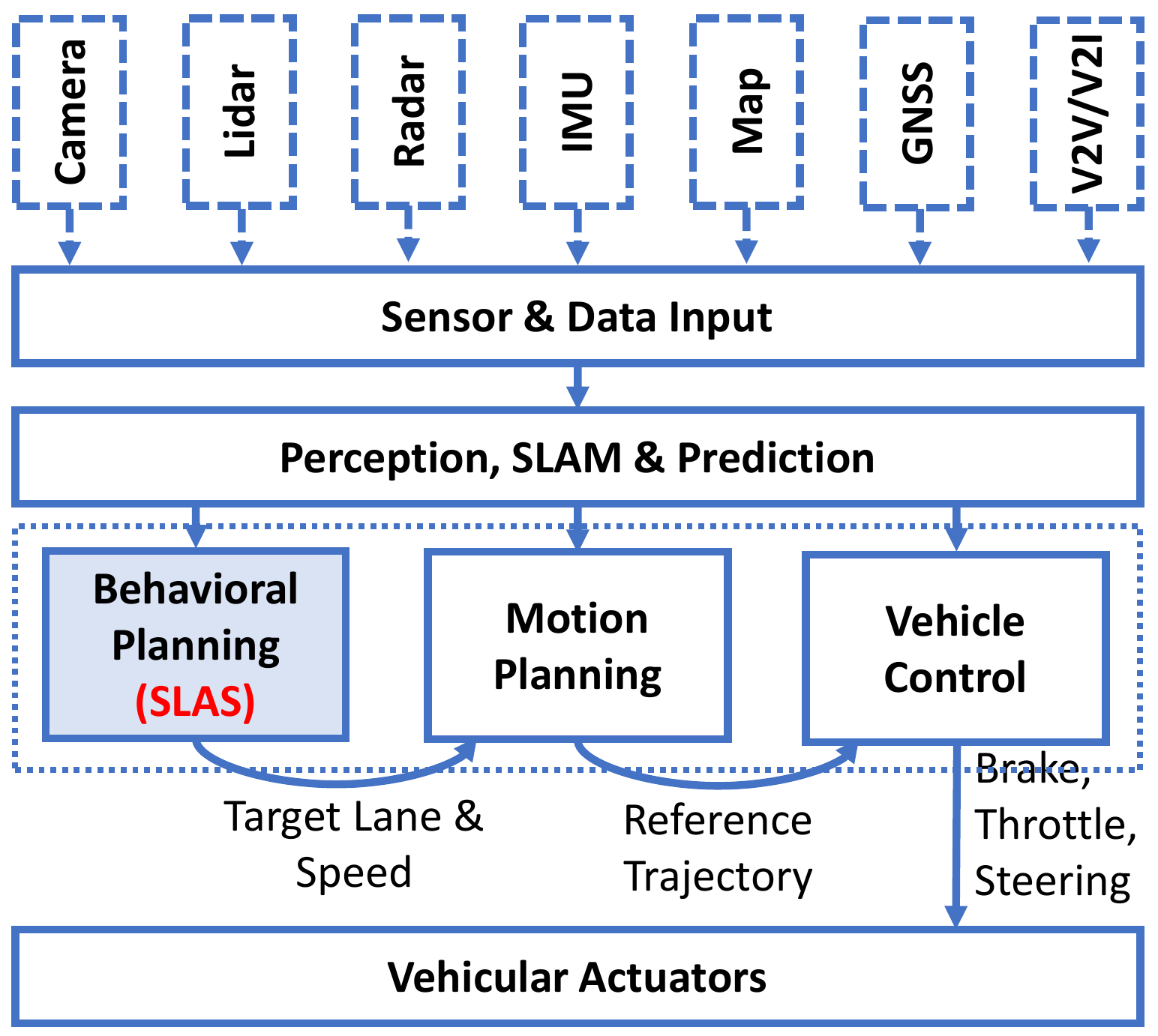}
\caption{Algorithmic pipeline of the proposed navigation architecture. The raw sensory input data is processed by the Perception, and Simultaneous Localization and Mapping (SLAM) modules to place the autonomous vehicle relative to the various environmental entities in a unified frame of reference. This information is then passed on to the navigation stack, composed of the behavioral planning, motion planning, and vehicle control modules. The output of the navigation module is passed down further in terms of actuation commands (brake, throttle and steering) to the actuators.}
\vspace{-10pt}
\label{fig:pipeline}
\end{figure}

Fig.~\ref{fig:pipeline} illustrates the algorithmic pipeline of the proposed navigation architecture, in reference to the various existing algorithmic modules deployed on an autonomous vehicle. The taxonomy of the various components of the navigation stack (highlighted by the dotted rectangle) is borrowed from \cite{surveyPlanning}. This pipeline essentially improves the pipeline introduced in \cite{bae2021risk} by adding a speed advisory system.

Our main focus is the development of the behavior planning module, highlighted as SLAS in Fig.~\ref{fig:pipeline}. SLAS outputs the target lane and reference speed which are utilized by the motion planning module to generate a reference trajectory for the ego vehicle. The vehicle controllers compute the throttle and steering commands to track the trajectory accordingly.



For the motion planning module, we adopt the Neural Networks integrated Model Predictive Control (NNMPC) \cite{nnmpc} due to its ability to accommodate the behaviors of neighboring vehicles in the trajectory generation process. In our approach, we assume that the perception (of other vehicles) and the localization (of ego vehicle) are known without any uncertainty, for simplicity, but the modular architecture avails us the ability to integrate any perception or SLAM module in the overall framework. 

\subsection*{Notation} 
Throughout the manuscript, $\mathbb{Z}$ will denote the set of integers and $\mathbb{R}$ the set of real numbers. For some $a,c \in \mathbb{Z}$ and $a<c$, we will write $\mathbb{Z}_{[a,c]} = \{b \in \mathbb{Z} \mid a \leq b \leq c \}$. For some $e,g \in \mathbb{R}$ and $e<g$, we will write $\mathbb{R}_{[e,g]} = \{f \in \mathbb{R} \mid e \leq f \leq g \}$. 

\subsection{Road Model}
The physical road structure is modeled as a continuous multi-lane highway with negligible curvature and unidirectional traffic flow. The lanes on the highway are clearly demarcated and at any given time $k$, the number of available lanes for the vehicles to travel on is denoted by $N_l(k)$ while the road speed limit is denoted by $V_l$. Therefore, the set of lanes available for traveling at a given time instant $k$ is denoted by $\mathbb{L}(k) = \mathbb{Z}_{[0,N_l(k)]}$. We work with the Frenet coordinate system where the distance along the road is denoted by the longitudinal displacement ($s$) and the distance perpendicular to the road is defined by the lateral displacement ($d$). Each lane is assigned a lane indicator variable $l$. The leftmost lane, with respect to the direction of traffic flow, is assigned a value of $l=0$ while each subsequent lane is assigned an increasing integer value for $l$, as depicted in Fig. \ref{fig:overview}.

\subsection{Vehicle Model}
Since we aim to have real-time computations for a long planning horizon ($> 15s$), we model the vehicle dynamics with a linearized decoupled dynamical system. For the highway driving scenario, where the road curvature is typically small, it is reasonable to assume a decoupling between the lateral and the longitudinal dynamics \cite{decoupledDynamics}, especially for the behavior planning layer. Therefore, we utilize a linear constant acceleration model for the longitudinal dynamics and abstract out the lateral dynamics with a lane indicator variable. For the lane change dynamics, we use a moving average filter coupled with a rounding function to model the time required by the ego vehicle to change lanes. This is compactly represented as:
\begin{align} \label{eqn:dynamicsLong}
s_0(k) &= s_0(k-1) + \frac{v_0(k-1) + v_0(k)}{2} \cdot T_s \\
l_0(k) &= \floor{\frac{1}{N} \sum_{i=0}^{N-1} \mathcal{L}(k-i) + \frac{1}{2}}
\end{align}
where $s_0(k)$, $v_0(k)$, $l_0(k)$ and $\mathcal{L}(k)$ denote the ego vehicle's longitudinal displacement, speed, lane indicator and target lane, respectively, at time instant $k$; the subscript $_i$ indexes the vehicles on the road with $_0$ being reserved for the ego vehicle; $T_s$ denotes the discretization time step; and $N$ corresponds to the number of time steps required to change lane. The state ($x_0(k)$) and control input ($u_0(k)$) to the system at time instant $k$ are defined as:
\begin{align}
    x_0(k) &= \begin{bmatrix}
    s_0(k) & l_0(k)
    \end{bmatrix} ^\T
    \in \mathbb{R} \times \mathbb{L}(k)
    \\
    u_0(k) &= \begin{bmatrix}
    v_0(k) & \mathcal{L}(k)
    \end{bmatrix} ^\T
    \in \mathbb{R}_{[0, V_{m}]} \times \mathbb{L}(k) \label{eqn:controlInput}
\end{align}
where $V_m$ denotes the maximum speed of the ego vehicle.

\subsection{Observation Model}
For practical considerations, we restrict the ego vehicle's visibility range to the sensory perception limit, denoted by $R_v$. Then, the set of vehicles in ego vehicle's visibility range at time instant $k$, represented by $\mathbb{O}(k)$, is defined as:
\begin{equation}
    \mathbb{O}(k) = \{i \in \mathbb{Z}_{> 0} \mid |s_i(k) - s_0(k)| \leq R_v\}
\end{equation}
where $s_i(k)$ corresponds to the longitudinal displacement of the observed vehicle.

\begin{remark}
    For the multi-lane highway driving scenario, occlusion does not play a prominent role so we do not account for it in the existing formulation. However, the proposed framework can easily accommodate occlusion and measurement uncertainties since the receding horizon approach bases its decision on the most up-to-date information available at any given time, as demonstrated in \cite{overtakingBidirectional}.
\end{remark}

\section{Methodology}
In this section, we describe the prediction model to generate the predicted future trajectories of observed vehicles and present a discussion on the proposed receding horizon optimization-based behavioral planning module.



\subsection{Trajectory Prediction}
\label{sec:pred}
Reliable behavior and trajectory prediction of other traffic participants is crucial for safe maneuvering of autonomous vehicles.
The algorithm proposed in Section \ref{sec:slas} is able to incorporate any generic prediction module available in the literature \cite{surveyPrediction} as long as it can provide a deterministic predicted future trajectory for a given vehicle. In this work, we formulate a low-complexity prediction model that highlights the flexibility and efficiency of our proposed approach.

For an observed vehicle $i \in \mathbb{O}(k)$, the future speed profile is predicted using a piece-wise linear function while the lane profile is assumed to stay constant for the duration of the prediction horizon. At a given time step $k$, the estimated acceleration ($\bar a_i^k$) and the estimated speed ($\bar v_i^k$) parameters are obtained through linear regression with mean-squared error on the past $o^k_i > 1$ speed observations. Based on the estimated parameters, we predict the future speed and longitudinal displacement as follows:
\begin{align}
\hat v_i^k(j) &= \begin{cases}
\bar v_i^k, & j=0 \\
\hat v_i^k(j-1) + \bar a_i^k \cdot j, & 0 < j \leq H_a \\
\hat v_i^k(j-1), & j > H_a
\end{cases} \label{eqn:velEstimation} \\
\hat s_i^k(j) &= \begin{cases}
s_i(k), & j=0 \\
\hat s_i^k(j-1) + \frac{T_s}{2} \cdot (\hat v_i^k(j-1) + \hat v_i^k(j)), & j>0. \label{eqn:posEstimation}
\end{cases}
\end{align}
Here, $H_a$ corresponds to the acceleration horizon while $\hat v_i^k(j)$ and $\hat s_i^k(j)$ respectively represent the predicted speed and longitudinal displacement for vehicle $i$, $j$ time steps into the future starting from the current time instant $k$.

\begin{remark}
    Due to the modular nature of the proposed framework, the behavior planning module detailed in Section \ref{sec:slas} can work with advanced maneuver-based (e.g. Markov Chain \cite{predManeuver}) and interaction-based (e.g. Social Generative Adversarial Networks \cite{predInteraction}) trajectory prediction modules, allowing for interactive maneuvering behaviors.
\end{remark}

\subsection{Speed and Lane Advisory System}
\label{sec:slas}
The goal of our behavior planning module, Speed and Lane Advisory System or in short, SLAS, is to determine a sequence of speed and lane change commands that would enable the ego vehicle to maximize its speed, thus minimizing the travel time, while accounting for driver comfort and abiding by its dynamical, actuator, and safety limits. The output of this module is a relatively smooth speed and lane change profile which is then passed on to a motion planner. It is necessary to incorporate the dynamical and actuator limits in the behavioral planning module so as not to provide the motion planner with goals that are not reachable, and jeopardize the safety of the overall system as a result.

In the subsequent discussion, we provide a formulation of the optimization problem for SLAS; highlight the modifications necessary to improve the computational complexity; and, present safety and feasibility analysis.

\subsubsection{Optimization Problem with Integer Constraints}
SLAS is posed as an optimization problem, with the objective to maximize speed while minimizing frequent lane changes and abrupt changes in speed. The output of SLAS, at time instant $k$, is the control input $u_0(k+1)$, as defined in (\ref{eqn:controlInput}). The optimization problem is formulated as follows:
\begin{align}
\min_{\substack{{v}^k(1), \cdots, {v}^k(H); \\ \mathcal{L}^k(1), \cdots, \mathcal{L}^k(H)}} \ & \sum_{j=1}^{H} \ [-\gamma_1 \cdot {v}^k(j) + \gamma_2 \cdot (\mathcal{L}^k(j) - \mathcal{L}^k(j-1))^2 \nonumber \\ & \qquad + \gamma_3 \cdot ({v}^k(j)-{v}^k(j-1))^2] \label{eqn:mimpcObj} \\
\textrm{s.t.} \quad & {s}^k(0) = 0 \label{eqn:mimpcDispInit}\\
& {v}^k(0) = v_0(k) \label{eqn:mimpcVelInit}\\
& \mathcal{L}^k(p) = l_0(k), \quad \forall p \in \mathbb{Z}_{[-N+1,0]} \label{eqn:mimpcLaneInit}\\
\forall j \in \, &\mathbb{Z}_{[1,H]}: \nonumber \\
& {v}^k(j) \in \mathbb{R}_{[0,V_{l}]} \label{eqn:mimpcVelLim}\\
& \frac{{v}^k(j) - {v}^k(j-1)}{Ts} \in \mathbb{R}_{[A_\text{min},A_\text{max}]} \label{eqn:mimpcAccelLim}\\
& {s}^k(j) = {s}^k(j-1) + \frac {{v}^k(j-1) + {v}^k(j)}{2} \cdot T_s \label{eqn:mimpcDisp}\\
& \mathcal{L}^k(j) \in \mathbb{L}(k) \label{eqn:mimpcLaneLim}\\
& \mathcal{L}^k(j) - \mathcal{L}^k(j-1) \in \mathbb{Z}_{[-1, 1]} \label{eqn:mimpcLaneChange}\\
& l^k(j) = \floor{ \frac{1}{N} \sum_{i=0}^{N-1} \mathcal{L}^k(j-i) + \frac{1}{2} } \label{eqn:mimpcLaneChangeTime}\\
& \min_{i \in \mathbb{A}(k)} \{|\hat {s}_i^k(j) - {s}^k(j) |\} \geq L_i^s(j), \label{eqn:mimpcSafety}\\
& \qquad \mathbb{A}(k) = \{a \in \mathbb{O}(k) \mid  l^k(j) = l_a(k)\} \nonumber.
\end{align}


\subsubsection*{Objective Function}
In the formulation above, the optimization variables are the ego vehicle's speed (${v}^k(j)$) and target lane ($\mathcal{L}^k(j)$), $j$ step into the future, starting from time instant $k$. Here, $H$ corresponds to the planning horizon. The scalarization parameters $\gamma_1$, $\gamma_2$ and $\gamma_3$ in the objective function (\ref{eqn:mimpcObj}) account for a relative tradeoff between maximizing speed, minimizing lane changes and minimizing abrupt changes in speed respectively. Increasing $\gamma_1$ yields a more aggressive behavior with the priority placed on maximizing speed while $\gamma_2$ and $\gamma_3$ combine to place an emphasis on maximizing passenger comfort by reducing lane and speed changes respectively.

\subsubsection*{Dynamical Constraints} These constraints are put in place to ensure the dynamical feasibility of the solution. The constraints (\ref{eqn:mimpcDispInit}), (\ref{eqn:mimpcVelInit}) and (\ref{eqn:mimpcLaneInit}) serve to initialize the longitudinal displacement, speed and target lane respectively for the optimizer, based on the values observed at time instant $k$. The constraints (\ref{eqn:mimpcVelLim}) and (\ref{eqn:mimpcAccelLim}) bound the ego vehicle's speed by the speed limit and the acceleration limits of the vehicle respectively. The ego vehicle's speed is then used to calculate the projected longitudinal displacement in (\ref{eqn:mimpcDisp}).

The target lane values at any planning step ($j$) are restricted to the set of reachable values by (\ref{eqn:mimpcLaneLim}), (\ref{eqn:mimpcLaneChange}) and (\ref{eqn:mimpcLaneChangeTime}). Here, (\ref{eqn:mimpcLaneLim}) restricts the target lane to the set of available lanes ($\mathbb{L}(k)$), (\ref{eqn:mimpcLaneChange}) ensures that the lane change, if needed, is made to the adjacent lane only and (\ref{eqn:mimpcLaneChangeTime}) models the time steps ($N$) required for a lane change. The flooring function can easily be transformed into a couple of linear constraints by the introduction of an auxiliary integer variable, as shown in the Appendix. Finally, $l^k(j)$ is merely the internal representation of the lane the ego vehicle is projected to travel on at planning step $j$.

\subsubsection*{Safety Constraint}
The safety constraint (\ref{eqn:mimpcSafety}) ensures that the ego vehicle maintains a minimum safe distance ($L^s_i(j)$) to the nearest vehicle $i$, in its projected lane of travel ($l^k(j)$), at planning instant $j$. We borrow the definition of this safe distance from \cite{pekFormalSafety}, where the authors provide a formalization, based on the clause from Vienna Convention on Road Traffic that states that ``\textit{A vehicle [...]  shall keep at a sufficient distance [...] to avoid collision if the vehicle in front should suddenly slow down or stop}." Furthermore, the absolute value constraint can be decomposed into linear constraints by the application of big-M method and the introduction of an auxiliary variable, as shown in the Appendix.

\begin{remark}
    The proposed formulation can accommodate arbitrary number of lanes at any given time instant $k$. This means that if at any given time, the number of available lanes for traveling either increases or decreases, the proposed formulation will still continue to hold. This is an important consideration since many a times on highways, some lanes are blocked due to various unanticipated situations such as road accidents, roadwork, narrowing of road etc.
\end{remark}

\subsubsection{Computational Complexity Reduction}
\label{sec:complexityReduction}
This section details the optimization problem reformulation with binary variables, optimization warm start technique and lazy constraint implementation, all of which combine to improve the computational complexity of our SLAS module.

\subsubsection*{Binary Variables}
The proposed formulation in Section \ref{sec:slas} has relatively high computation complexity (computation time of $\sim 2s$  in the worst case scenario - slow moving traffic blocking all the lanes) due to the integer decision variables yielding a mixed-integer optimization problem \cite{mip}. To circumvent the computational overload, we reformulate the problem with binary variables that replace the integer variables, as follows:
\begin{equation}
    \forall i \in \mathbb{L}(k), \forall j \in \mathbb{Z}_{[1,H]}: \tilde {\mathcal{L}}^k(i,j) \in \{0,1\}
\end{equation}
where the $\tilde {\mathcal{L}}^k(i,j)$ represents the modified target lane variable, indexed by the lane ($i$) as well as the planning step ($j$) and $\tilde {\mathcal{L}}^k(a,b)=1$ represents the choice of lane $a \in \mathbb{L}$ as the target lane at planning step $b \in \mathbb{Z}_{[1,H]}$. Then, some of the constraints from the SLAS formulation in Section \ref{sec:slas} are modified as follows:
\begin{align}
    (\ref{eqn:mimpcLaneInit}) &\rightarrow \tilde {\mathcal{L}}^k(l_0(k),0) = 1 \label{eqn:mbmpcLaneInit}\\
    (\ref{eqn:mimpcLaneLim}) &\rightarrow \sum_{i \in \mathbb{L}(k)} \tilde {\mathcal{L}}^k(i,j) = 1, \forall j \in \mathbb{Z}_{[1,H]} \label{eqn:mbmpcLaneLim}\\
     (\ref{eqn:mimpcLaneChange}) &\rightarrow \tilde {\mathcal{L}}^k(a,j) = 1 \implies \sum_{b \in \mathbb{B}(k)} \tilde {\mathcal{L}}^k(b,j) = 1, \label{eqn:mbmpcLaneChange}\\
     & \qquad \qquad \mathbb{B}(k) = \mathbb{Z}_{[a-1,a+1]} \cap \mathbb{L}(k) \nonumber \\
     (\ref{eqn:mimpcSafety}) &\rightarrow \min_{i \in \mathbb{A}(k)} \{|\hat {s}_i^k(j) - {s}^k(j) |\} \geq \tilde{L}_i^s(j), \label{eqn:mbmpcSafety} \\
& \qquad \mathbb{A}(k) = \{a \in \mathbb{O}(k) \mid \tilde {\mathcal{L}}^k(l_a(k),j) = 1 \} \nonumber.
\end{align}
Here, (\ref{eqn:mbmpcLaneInit}) initializes the target lane, (\ref{eqn:mbmpcLaneLim}) restricts the target lane at any planning step to the set of available lanes, (\ref{eqn:mbmpcLaneChange}) restricts the lane change between consecutive planning steps to the adjacent lanes, and (\ref{eqn:mbmpcSafety}) represents the augmented safety constraint. The implication ($\implies$) in (\ref{eqn:mbmpcLaneChange}) can easily be transformed into a linear constraint (see Appendix). The augmented minimum safety distance ($\tilde{L}_i^s(j)$) incorporates the time required to execute the lane change maneuver ($N$) from (\ref{eqn:mimpcLaneChangeTime}) into the following unified safety constraint:
\begin{align}
    \tilde{L}^s_i(j) &= L^s_i(j) + \gamma_d (\delta^k (j)) \cdot (v^k(j) \cdot N \cdot T_s)\\
    \gamma_d (\delta^k (j)) &= \gamma_4 \cdot \frac{2|\delta^k(j)|}{L_l}\\
    \delta^k (j) &= \min\left\{\delta(k) + \frac{L_l \cdot j}{N}, \frac{L_l}{2} \right\}
\end{align}
where $L_l$ is the width of the lanes (see Fig. \ref{fig:overview}), $\delta(k)$ is the signed lateral deviation of the ego vehicle from the \textbf{previous} target lane's boundary at time step $k$, and $\gamma_d(\delta^k(j))$ is the dynamic cost of deviation from the previous target lane ($\mathcal{L}(k-1)$). Moreover, in the cost function (\ref{eqn:mimpcObj}), we take $\mathcal{L}^k(0) = \mathcal{L}(k-1)$. These costs are introduced to prevent the swerving (canceling of lane switch before completion) behavior, unless absolutely necessary (for safety purposes).

\begin{remark}
    Since the ego vehicle is considered to have changed lane once it crosses a lane boundary, the deviation $\delta^k(j)$ is considered from the lane boundary instead of the center of the target lane to maintain the continuity of $\gamma_d(\delta^k(j))$ with respect to the lateral displacement of the ego vehicle.
    Specifically, $\delta^k(j) > 0$ if the ego vehicle has crossed the previous target lane boundary and $0$ otherwise.
    This is an important consideration since a discontinuity in $\gamma_d(\delta^k(j))$, upon completion of lane change, may lead to infeasibility.
\end{remark}

\begin{remark}
    The swerving behavior is suppressed but not completely eliminated with a hard constraint since such a behavior is necessary at times to react to the environment's unpredictability. This reactive strategy, which is a distinctive feature of our approach, avails the algorithm the ability to proactively `change its mind' in case something unanticipated happens in the environment that can jeopardize safety.
\end{remark}

\subsubsection*{Optimization Warm Start}
To aid the optimizer in finding an initially feasible solution, we provide the solution from the previous time step as a reference. Formally, $\{v^{k-1}(2), \cdots, v^{k-1}(H), \mathcal{L}^{k-1}(2), \cdots, \mathcal{L}^{k}(H)\}$ is provided as a reference for $\{v^{k}(1), \cdots, v^{k}(H-1), \mathcal{L}^{k}(1), \cdots, \mathcal{L}^{k}(H-1)\}$. This doesn't imply that the solution from time step $k-1$ will hold exactly at time step $k$, owing to the unmodeled disturbances, but providing this reference aids the optimizer in finding an initially feasible solution in the vicinity of the reference solution. This observation is rooted in the premise that the solution for the long planning horizon is not expected to change significantly between time steps, given the sampling time is not too large, and the predicted behavior of on-road agents does not alter significantly.

It is also worth pointing out that the priority here is quickly finding a feasible solution that obeys the safety constraints and actuator limits, and recursively improving it rather than excessively iterating to reach at a global optimum. In our experiments, it was observed that a suboptimal solution was qualitatively not significantly different from the optimal one. Therefore, we utilize the cutting planes method for optimization \cite{cuttingPlane}, which first looks for a feasible solution, using our provided reference, and then recursively updates it until either the globally minimal solution is found or the time limit is reached.

\subsubsection*{Lazy Constraints}
To further enhance the computational efficiency, we introduce a lazy implementation of the lane changing constraints (\ref{eqn:mbmpcLaneChange}). It was observed in our experiments that a feasible solution without the lane changing constraints (\ref{eqn:mbmpcLaneChange}) can be found several order of magnitude ($\sim 10\times$) quicker than if we include these constraints so we decided to have a lazy implementation for them. With a lazy implementation \cite{lazy}, the solver finds a set of feasible solutions without the inclusion of these constraints and then determines the feasibility of those solutions from the reduced problem with respect to the lazy constraints.

\subsubsection{Feasibility}
By an argument similar to the one presented in \cite{overtakingBidirectional}, it is a relatively straightforward proof for recursive feasibility of the problem, i.e. the optimization problem will continue to stay feasible, if initially feasible, with the trivial solution being matching the speed of the leading vehicle and not changing lanes.

\section{Results}
\label{sec:results}
In this section, we detail our experimental setup, demonstrate the performance of SLAS, and report a qualitative as well as a quantitative comparative analysis. The baselines in our comparative analysis are set to: Extended-Astar (EA$^\star$) \cite{bae2021risk}, MOBIL \cite{mobil}, and no lane-change model (No-Change).

\subsection{Experimental Setup}
\begin{figure} [ht]
\centering
\includegraphics[trim=0 0 0 0, clip,width=0.9\columnwidth]{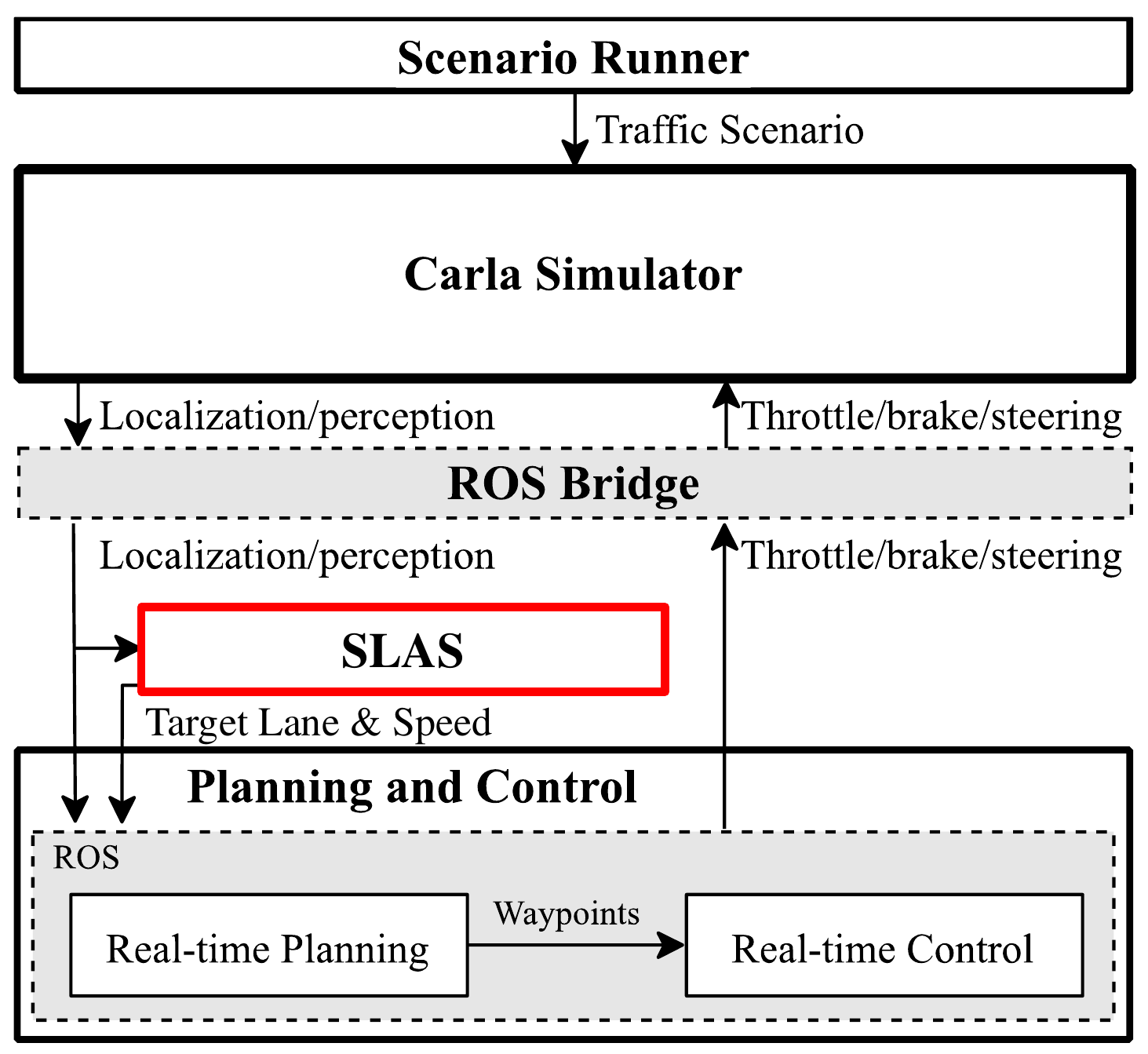}
\caption{Simulation Setup. Scenario Runner sets up the scenario for the CARLA Simulator, which then communicates with the SLAS and the Planning and Control ROS (Robot Operating System) nodes through the ROS bridge node.}
\label{fig:setup}
\vspace{-15pt}
\end{figure}

The implementation setup, depicted in Fig. \ref{fig:setup}, is composed of the CARLA simulator (Version 0.9.11) \cite{carla}, SLAS module (Section \ref{sec:slas}), and the planner and controller module \cite{nnmpc}. To solve the optimization problem for SLAS, we use Gurobi Optimizer (Version 9.1.1) \cite{gurobi}. The simulations are performed on a computer equipped with an Intel Xeon(R) CPU E5-2643 v4 @ 3.40GHz × 12 and NVIDIA Titan XP, running Ubuntu 20.04 LTS. On average, the time required for each optimization step is $\sim 0.096s$, while the maximum time limit for the optimizer is set to $0.2s$, indicating the strong potential for real-time applicability. 

\subsection{Case Study}
\label{sec:caseStudy}
\begin{figure*} [ht]
\centering
\includegraphics[trim=0 0 0 0, clip,width=0.95\textwidth]{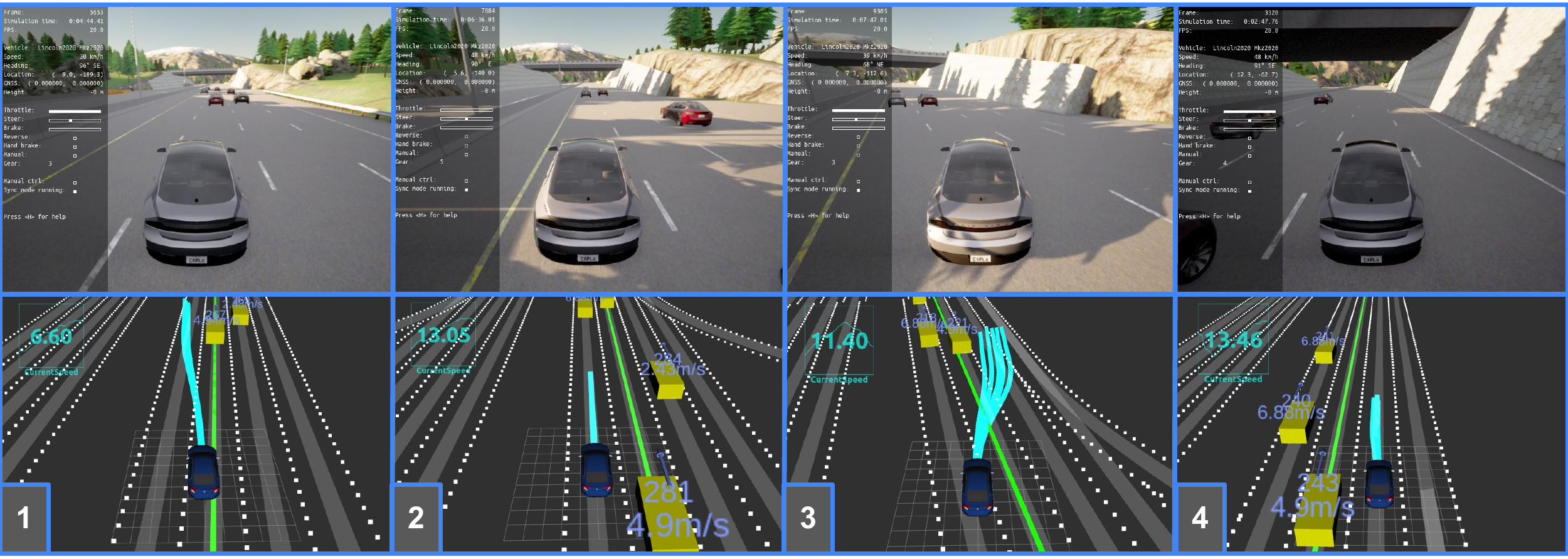}
\caption{Testing scenario with three lanes: lane 0 (left), lane 1 (center) and lane 2 (right). The expected motion of the ego vehicle, over the course of the simulation, is shown with numbered frames. The right most lane (lane 3) is reserved for merging traffic so it is not utilized in our simulation.}
\label{fig:scenario}
\end{figure*}
Figure~\ref{fig:scenario} illustrates the test case scenario for our comparative analysis. The scenario is composed of a highway segment with four lanes and the rightmost lane reserved for merging vehicles.
The ego vehicle is initialized to follow a slow moving vehicle in lane 1 and has even slower moving traffic to its right in lane 2. Thus, the only option for it, in order to minimize travel time, is to switch left to lane 0 with faster moving traffic and greater headway. Once it moves to lane 0, and overtakes the slow moving vehicle in lane 1, it has two options: either to keep traveling in lane 0 without making any lane change decisions until getting close to the lead vehicle or proactively exploiting the gap in lane 2 to switch to lane 3 in anticipation of traffic buildup in lanes 1 and 2. A strategic decision maker with foresight will choose to take the later option and make the decision proactively for a greater overall benefit.

The evaluation metrics for the comparative analysis include: travel time, lateral displacement, headway and distance to the closest vehicle. As for the simulation parameters, the simulation step size is set to $0.05s$ (simulation frequency of $20Hz$); the velocities of vehicles in lanes 0, 1 and 2 are set to 8, 5 and 2 $m/s$ respectively while the speed limit $V_l$ is set to $15 m/s$; the length of the highway patch is set to $350m$ while the width between the lanes is set to $3.5m$; and the sensor visibility range is set to $R_v = 50m$. The parameters for SLAS are set as follows: $T_s = 0.4s$, $H=40$, $N=3$, $A_{min} = -5 m/s^2$, $A_{max}=3.5 m/s^2$,  $\gamma_1 = 1$, $\gamma_2 = 0.1$ and $\gamma_3 = 0.01$. The values of these parameters can be tuned to yield an aggressive or defensive behavior of the algorithm.

\begin{figure*}
    \centering
    \includegraphics[width=0.65\columnwidth]{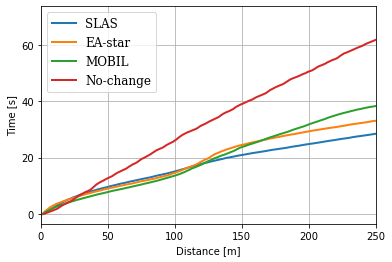}\label{fig:time}
    \includegraphics[width=0.65\columnwidth]{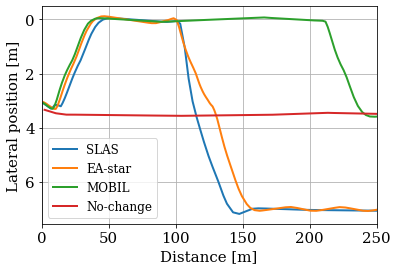}\label{fig:lateral}
    \includegraphics[width=0.65\columnwidth]{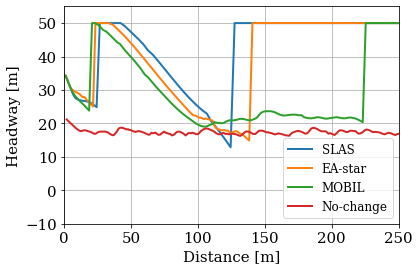}\label{fig:headway}
    \caption{Left: Travel time comparison. Center: Lane choice (lateral position) comparison. The center lines of lanes 0 (left), 1 (center) and 2 (right) have fixed lateral displacements of $0m$, $3.5m$ and $7m$ respectively. Right: Headway comparison. With no leading vehicle, the headway is restricted by the visibility range of $50m$.}
    \label{fig:trajectory}
    \vspace{-10pt}
\end{figure*}

\subsubsection{Travel Time}

The left plot in Fig.~\ref{fig:trajectory} depicts the travel time as a function of longitudinal displacement for the four algorithms. As seen in the plot, our method (SLAS) maintains a lower overall travel time as compared to the other methods. Quantitatively speaking, SLAS outperforms EA$^\star$, MOBIL and No-change methods by $12.72\%$, $23.52\%$ and $54.34\%$ respectively in terms of the time required to complete the simulation scenario. This shows that our method's foresight compensates for its apparent conservativeness arising from the need to preserve passenger comfort.

\subsubsection{Lateral Displacement}

To identify the differences in lane changing behaviors between the four approaches, the relationship between lateral and longitudinal displacements over the course of the simulation is highlighted in the center plot of Fig. \ref{fig:trajectory}. In the plot, the lateral displacement of 0 corresponds to the center of lane 0 while the center of each following lane is $3.5m$ away. Comparing the performance of the four algorithms, we see SLAS and EA$^\star$ showing relatively similar performances, resulting from proactive decision making. In contrast, since MOBIL only assesses the advantage of switching to the adjacent lanes, it is unable to see the benefit of proactively switching to lane 2. This explains why EA$^\star$ and SLAS start outperforming MOBIL in terms of travel time (left plot) at around the 130 [m] mark for longitudinal displacement.

As for a direct comparison between SLAS and EA$^\star$, the benefits of having speed advisory system become apparent in this center plot. Due to speed control, SLAS 
is able to constantly maintain a greater headway (right plot) without having to brake significantly upon getting too close to the lead vehicle. This results in a smooth lateral displacement profile which allows the vehicle to change lanes with minimal jerk (quantitative analysis to follow in Section \ref{sec:monteCarlo}) and deliver better overall timing performance (left plot).

\subsubsection{Headway}

The right plot in Fig. \ref{fig:trajectory} shows the headway maintained by the ego vehicle over the course of the simulation. In accordance with our prior discussion, MOBIL cruises behind the front vehicle, maintaining a relatively low headway until a sufficient space in the adjacent lane is found to perform the lane-change maneuver. 
On the other hand, EA$^\star$ and SLAS show a comparable headway trajectory, however, SLAS maintains a greater headway throughout and achieves the maximum headway prior to EA$^\star$. Quantitatively, SLAS maintains on average $9.43\%$, $36.57\%$ and $113.17\%$ more headway than the EA$^\star$, MOBIL and No-change approaches respectively. This strong performance by SLAS can be attributed to its incorporation of safety guarantees coupled with its consideration for passenger comfort.

\subsubsection{Distance to closest vehicle}
Finally, we compare the distance that ego vehicle maintains from the closest vehicle throughout the simulation. On average, SLAS maintains $9.28\%$, $32.01\%$, and $22.84\%$ more distance in comparison to EA$^\star$, MOBIL and No-change approaches respectively. These numbers are a testament to the strength of our approach resulting from consideration of long planning horizon coupled with speed control.

\subsection{Monte Carlo Simulations}
\label{sec:monteCarlo}

\begin{table}[ht]
\begin{tabular}{|cccccccc|}
\hline
\textbf{Model} &
  \textbf{\begin{tabular}[c]{@{}c@{}}Comp.\\ Time\end{tabular}} &
  \textbf{Brake} &
  \textbf{\begin{tabular}[c]{@{}c@{}}Brake\\ Jerk\end{tabular}} &
  \textbf{Thr.} &
  \textbf{\begin{tabular}[c]{@{}c@{}}Thr.\\ Jerk\end{tabular}} &
  \textbf{\begin{tabular}[c]{@{}c@{}}Ang.\\ Acc.\end{tabular}} &
  \textbf{\begin{tabular}[c]{@{}c@{}}Ang.\\ Jerk\end{tabular}} \\ \hline
\multicolumn{8}{|c|}{\textbf{Average}} \\ \hline
\multicolumn{1}{|c|}{\textbf{SLAS}} &
  \cellcolor[HTML]{FFFFFF}27.84 &
  \cellcolor[HTML]{34FF34}-0.46 &
  \cellcolor[HTML]{34FF34}-0.45 &
  \cellcolor[HTML]{34FF34}0.74 &
  \cellcolor[HTML]{34FF34}0.69 &
  \cellcolor[HTML]{34FF34}2.21 &
  \cellcolor[HTML]{34FF34}5.04 \\
\multicolumn{1}{|c|}{EAstar} &
  \cellcolor[HTML]{34FF34}27.23 &
  \cellcolor[HTML]{FFFFFF}-0.56 &
  \cellcolor[HTML]{FFFFFF}-0.47 &
  \cellcolor[HTML]{FFFFFF}0.83 &
  \cellcolor[HTML]{FFFFFF}0.77 &
  \cellcolor[HTML]{FFFFFF}2.83 &
  \cellcolor[HTML]{FFFFFF}6.69 \\
\multicolumn{1}{|c|}{MOBIL} &
  \cellcolor[HTML]{FFFFFF}28.06 &
  \cellcolor[HTML]{FFFFFF}-0.62 &
  \cellcolor[HTML]{FFFFFF}-0.49 &
  \cellcolor[HTML]{FFFFFF}0.86 &
  \cellcolor[HTML]{FFFFFF}0.79 &
  \cellcolor[HTML]{FFFFFF}2.43 &
  \cellcolor[HTML]{FFFFFF}5.62 \\ \hline
\multicolumn{8}{|c|}{\textbf{Standard Deviation}} \\ \hline
\multicolumn{1}{|c|}{\textbf{SLAS}} &
  \cellcolor[HTML]{34FF34}1.77 &
  \cellcolor[HTML]{34FF34}0.25 &
  \cellcolor[HTML]{34FF34}0.09 &
  \cellcolor[HTML]{34FF34}0.08 &
  \cellcolor[HTML]{34FF34}0.15 &
  \cellcolor[HTML]{34FF34}0.73 &
  \cellcolor[HTML]{34FF34}1.89 \\
\multicolumn{1}{|c|}{EAstar} &
  \cellcolor[HTML]{FFFFFF}2.85 &
  \cellcolor[HTML]{FFFFFF}0.38 &
  \cellcolor[HTML]{FFFFFF}0.15 &
  \cellcolor[HTML]{FFFFFF}0.11 &
  \cellcolor[HTML]{FFFFFF}0.21 &
  \cellcolor[HTML]{FFFFFF}1.48 &
  \cellcolor[HTML]{FFFFFF}3.93 \\
\multicolumn{1}{|c|}{MOBIL} &
  \cellcolor[HTML]{FFFFFF}3.82 &
  \cellcolor[HTML]{FFFFFF}0.42 &
  \cellcolor[HTML]{FFFFFF}0.18 &
  \cellcolor[HTML]{FFFFFF}0.11 &
  \cellcolor[HTML]{FFFFFF}0.23 &
  \cellcolor[HTML]{FFFFFF}0.77 &
  \cellcolor[HTML]{FFFFFF}2.05 \\ \hline
\end{tabular}
\caption{Monte Carlo Simulation Results}
\label{tab:monteCarlo}
\vspace{-15pt}
\end{table}

To demonstrate the long-term performance of the three approaches (SLAS, EA$^\star$ and MOBIL), we run a series of Monte Carlo simulations on scenarios with randomized initial positions (within a range of $8m$) and velocities (within ranges of $8$, $5$ and $2$ $m/s$ assigned to each of the three lanes randomly) of traffic participants. The result from 50 simulations is presented in Table \ref{tab:monteCarlo}.

In this table, the columns represent the different evaluation metrics, the rows identify the three algorithms, and the values highlighted in green represent the best result with respect to each evaluation metric. The evaluation metrics, going from left to right in the table, are completion time ($s$), brake ($\mathbb{R}_{[-1,0]}$), brake jerk ($\mathbb{R}_{[-1,0]}$), throttle ($\mathbb{R}_{[0,1]}$), throttle jerk ($\mathbb{R}_{[0,1]}$), angular acceleration ($^{\circ}/s^2$) and angular jerk ($^{\circ}/s^3$). Apart from completion time, the remaining metrics, based on the commands passed to the vehicular actuators (Fig. \ref{fig:pipeline}), are used to model passenger comfort. In terms of average performance, SLAS greatly outperforms the other methods when it comes to passenger comfort since it explicitly accounts for comfort in the formulation. However, it does so at a cost of slightly reduced performance in regards to travel time, when compared to EA$^\star$, since SLAS tries to strike a balance between minimizing travel time and maximizing passenger comfort.
SLAS also secures the lowest standard deviation, for each of the evaluation metrics, when compared to the other methods, which points to the consistency in its long-term performance.

\section{Conclusion}
\label{sec:conclusion}
We propose a novel behavior planning module for the multi-lane highway maneuvering scenario, that outputs strategic target lane and reference speed commands, and incorporate it with a state-of-the-art motion planning and control framework. We formulate the approach as a receding horizon mixed integer optimization with the goal to minimize travel time while accounting for passenger comfort for a long planning horizon. In order to reduce the computational overload, we reformulate the problem by replacing integer variables with binary ones and further incorporate various modifications to aid numerical computations. We also carry out a detailed comparative analysis to demonstrate the performance of our approach on the CARLA simulator. Our future work includes incorporating various delays and uncertainty measures in the perception, localization and prediction modules to evaluate the robustness properties of our approach.


\typeout{}
\bibliographystyle{IEEEtran}
\bibliography{references}

\begin{thebibliography}{10}
\providecommand{\url}[1]{#1}
\csname url@samestyle\endcsname
\providecommand{\newblock}{\relax}
\providecommand{\bibinfo}[2]{#2}
\providecommand{\BIBentrySTDinterwordspacing}{\spaceskip=0pt\relax}
\providecommand{\BIBentryALTinterwordstretchfactor}{4}
\providecommand{\BIBentryALTinterwordspacing}{\spaceskip=\fontdimen2\font plus
\BIBentryALTinterwordstretchfactor\fontdimen3\font minus
  \fontdimen4\font\relax}
\providecommand{\BIBforeignlanguage}[2]{{%
\expandafter\ifx\csname l@#1\endcsname\relax
\typeout{** WARNING: IEEEtran.bst: No hyphenation pattern has been}%
\typeout{** loaded for the language `#1'. Using the pattern for}%
\typeout{** the default language instead.}%
\else
\language=\csname l@#1\endcsname
\fi
#2}}
\providecommand{\BIBdecl}{\relax}
\BIBdecl

\bibitem{laneChangingRisk}
G.~Liu, S.~Chen, Z.~Zeng, H.~Cui, Y.~Fang, D.~Gu, Z.~Yin, and Z.~Wang, ``Risk
  factors for extremely serious road accidents: Results from national road
  accident statistical annual report of china,'' \emph{PLoS one}, vol.~13,
  no.~8, p. e0201587, 2018.

\bibitem{bae2021risk}
S.~Bae, D.~Isele, K.~Fujimura, and S.~J. Moura, ``Risk-aware lane selection on
  highway with dynamic obstacles,'' in \emph{2021 IEEE Intelligent Vehicles
  Symposium (IV)}.\hskip 1em plus 0.5em minus 0.4em\relax IEEE, 2021, pp.
  652--659.

\bibitem{mobil}
\BIBentryALTinterwordspacing
A.~Kesting, M.~Treiber, and D.~Helbing, ``General lane-changing model mobil for
  car-following models,'' \emph{Transportation Research Record}, vol. 1999,
  no.~1, pp. 86--94, 2007. [Online]. Available:
  \url{https://doi.org/10.3141/1999-10}
\BIBentrySTDinterwordspacing

\bibitem{motionPlanningLaneVelocity}
Y.~Chen, C.~Hu, and J.~Wang, ``Motion planning with velocity prediction and
  composite nonlinear feedback tracking control for lane-change strategy of
  autonomous vehicles,'' \emph{IEEE Transactions on Intelligent Vehicles},
  vol.~5, no.~1, pp. 63--74, 2019.

\bibitem{rrt}
W.~Khaksar, K.~S.~M. Sahari, and T.~S. Hong, ``Application of sampling-based
  motion planning algorithms in autonomous vehicle navigation,''
  \emph{Autonomous Vehicle}, vol. 735, 2016.

\bibitem{rrtFrazzoli}
Y.~Kuwata, J.~Teo, G.~Fiore, S.~Karaman, E.~Frazzoli, and J.~P. How,
  ``Real-time motion planning with applications to autonomous urban driving,''
  \emph{IEEE Transactions on control systems technology}, vol.~17, no.~5, pp.
  1105--1118, 2009.

\bibitem{mukadam2017tactical}
M.~Mukadam, A.~Cosgun, A.~Nakhaei, and K.~Fujimura, ``Tactical decision making
  for lane changing with deep reinforcement learning,'' 2017.

\bibitem{dqn}
C.-J. Hoel, K.~Wolff, and L.~Laine, ``Automated speed and lane change decision
  making using deep reinforcement learning,'' in \emph{2018 21st International
  Conference on Intelligent Transportation Systems (ITSC)}.\hskip 1em plus
  0.5em minus 0.4em\relax IEEE, 2018, pp. 2148--2155.

\bibitem{drl}
M.~{Kaushik}, V.~{Prasad}, K.~M. {Krishna}, and B.~{Ravindran}, ``Overtaking
  maneuvers in simulated highway driving using deep reinforcement learning,''
  in \emph{2018 IEEE Intelligent Vehicles Symposium (IV)}, 2018, pp.
  1885--1890.

\bibitem{yang2020cm3}
J.~Yang, A.~Nakhaei, D.~Isele, K.~Fujimura, and H.~Zha, ``Cm3: Cooperative
  multi-goal multi-stage multi-agent reinforcement learning,'' in
  \emph{International Conference on Learning Representations}, 2020.

\bibitem{saxena2020driving}
D.~M. Saxena, S.~Bae, A.~Nakhaei, K.~Fujimura, and M.~Likhachev, ``Driving in
  dense traffic with model-free reinforcement learning,'' in \emph{2020 IEEE
  International Conference on Robotics and Automation (ICRA)}.\hskip 1em plus
  0.5em minus 0.4em\relax IEEE, 2020, pp. 5385--5392.

\bibitem{potential}
S.~Glaser, B.~Vanholme, S.~Mammar, D.~Gruyer, and L.~Nouvelière,
  ``Maneuver-based trajectory planning for highly autonomous vehicles on real
  road with traffic and driver interaction,'' \emph{IEEE Transactions on
  Intelligent Transportation Systems}, vol.~11, no.~3, pp. 589--606, 2010.

\bibitem{optimalControl}
G.~Franze and W.~Lucia, ``A receding horizon control strategy for autonomous
  vehicles in dynamic environments,'' \emph{IEEE Transactions on Control
  Systems Technology}, vol.~24, no.~2, pp. 695--702, 2015.

\bibitem{mpcDriving}
P.~Falcone, F.~Borrelli, J.~Asgari, H.~E. Tseng, and D.~Hrovat, ``Predictive
  active steering control for autonomous vehicle systems,'' \emph{IEEE
  Transactions on Control Systems Technology}, vol.~15, no.~3, pp. 566--580,
  2007.

\bibitem{mpcPlanning}
C.~Liu, S.~Lee, S.~Varnhagen, and H.~E. Tseng, ``Path planning for autonomous
  vehicles using model predictive control,'' in \emph{2017 IEEE Intelligent
  Vehicles Symposium (IV)}.\hskip 1em plus 0.5em minus 0.4em\relax IEEE, 2017,
  pp. 174--179.

\bibitem{scenarioMPC}
G.~Schildbach and F.~Borrelli, ``Scenario model predictive control for lane
  change assistance on highways,'' in \emph{2015 IEEE Intelligent Vehicles
  Symposium (IV)}.\hskip 1em plus 0.5em minus 0.4em\relax IEEE, 2015, pp.
  611--616.

\bibitem{surveyPlanning}
B.~Paden, M.~{\v{C}}{\'a}p, S.~Z. Yong, D.~Yershov, and E.~Frazzoli, ``A survey
  of motion planning and control techniques for self-driving urban vehicles,''
  \emph{IEEE Transactions on intelligent vehicles}, vol.~1, no.~1, pp. 33--55,
  2016.

\bibitem{nnmpc}
S.~Bae, D.~Saxena, A.~Nakhaei, C.~Choi, K.~Fujimura, and S.~Moura,
  ``Cooperation-aware lane change maneuver in dense traffic based on model
  predictive control with recurrent neural network,'' in \emph{2020 American
  Control Conference (ACC)}.\hskip 1em plus 0.5em minus 0.4em\relax IEEE, 2020,
  pp. 1209--1216.

\bibitem{decoupledDynamics}
R.~Attia, R.~Orjuela, and M.~Basset, ``Combined longitudinal and lateral
  control for automated vehicle guidance,'' \emph{Vehicle System Dynamics},
  vol.~52, no.~2, pp. 261--279, 2014.

\bibitem{overtakingBidirectional}
F.~M. Tariq, N.~Suriyarachchi, C.~Mavridis, and J.~S. Baras, ``Autonomous
  vehicle overtaking in a bidirectional mixed-traffic setting,'' in \emph{2022
  American Control Conference (ACC)}, 2022.

\bibitem{surveyPrediction}
S.~Lef{\`e}vre, D.~Vasquez, and C.~Laugier, ``A survey on motion prediction and
  risk assessment for intelligent vehicles,'' \emph{ROBOMECH journal}, vol.~1,
  no.~1, pp. 1--14, 2014.

\bibitem{predManeuver}
K.~Okamoto, K.~Berntorp, and S.~Di~Cairano, ``Similarity-based vehicle-motion
  prediction,'' in \emph{2017 American Control Conference (ACC)}.\hskip 1em
  plus 0.5em minus 0.4em\relax IEEE, 2017, pp. 303--308.

\bibitem{predInteraction}
A.~Gupta, J.~Johnson, L.~Fei-Fei, S.~Savarese, and A.~Alahi, ``Social gan:
  Socially acceptable trajectories with generative adversarial networks,'' in
  \emph{Proceedings of the IEEE conference on computer vision and pattern
  recognition}, 2018, pp. 2255--2264.

\bibitem{pekFormalSafety}
C.~Pek, P.~Zahn, and M.~Althoff, ``Verifying the safety of lane change
  maneuvers of self-driving vehicles based on formalized traffic rules,'' in
  \emph{Proc. of the IEEE Intelligent Vehicles Symposium}, 2017.

\bibitem{mip}
A.~Richards and J.~How, ``Mixed-integer programming for control,'' in
  \emph{Proceedings of the 2005, American Control Conference, 2005.}, 2005, pp.
  2676--2683 vol. 4.

\bibitem{cuttingPlane}
H.~Marchand, A.~Martin, R.~Weismantel, and L.~Wolsey, ``Cutting planes in
  integer and mixed integer programming,'' \emph{Discrete Applied Mathematics},
  vol. 123, no. 1-3, pp. 397--446, 2002.

\bibitem{lazy}
B.~Cuteri, C.~Dodaro, F.~Ricca, and P.~Sch{\"u}ller, ``Constraints, lazy
  constraints, or propagators in asp solving: An empirical analysis,''
  \emph{Theory and Practice of Logic Programming}, vol.~17, no. 5-6, pp.
  780--799, 2017.

\bibitem{carla}
A.~Dosovitskiy, G.~Ros, F.~Codevilla, A.~Lopez, and V.~Koltun, ``Carla: An open
  urban driving simulator,'' in \emph{Conference on robot learning}.\hskip 1em
  plus 0.5em minus 0.4em\relax PMLR, 2017, pp. 1--16.

\bibitem{gurobi}
\BIBentryALTinterwordspacing
L.~Gurobi~Optimization, ``Gurobi optimizer reference manual,'' 2021. [Online].
  Available: \url{http://www.gurobi.com}
\BIBentrySTDinterwordspacing

\end{thebibliography}

\appendix
\subsection{Flooring Constraint}
For $y \in \mathbb{Z}$ and $x \in \mathbb{R}$, the constraint $y = \floor{x}$
can be represented by the following linear constraints:
\begin{equation*}
    y \leq x, \quad y + 1 \geq x + \epsilon
\end{equation*}
where $\epsilon > 0$ accounts for the feasibility tolerance.

\subsection{Implication Constraint}
For $a,b \in \{0,1\}$, the constraint $(a=1) \implies (b=1)$ can be represented with a linear constraint as follows:
\begin{equation*}
    b + M \cdot (1-a) \geq 1 - \epsilon
\end{equation*}
where $M\gg0$ (big-M) and $\epsilon > 0$ accounts for numerical errors (chosen to be $0.1$ in our implementations).

\subsection{Absolute Value Constraint} 
For $\Delta s, L_s \in \mathbb{R}$, the constraint $|\Delta s|-L_s \geq 0$ can be represented as: $$\Delta s \geq L_s \vee \Delta s \leq -L_s.$$ This can further be generalized, as done in our implementation, to have different forward and rear safety margins as:
$$\Delta s \geq L^f_s \vee \Delta s \leq -L^r_s$$
where $L^f_s$ and $L^r_s$ are forward and rear safety margins respectively. This can be represented with the following linear constraints:
\begin{align*}
    \Delta s + M \cdot c - L^f_s &\geq 0 \\
    -\Delta s + M \cdot (1-c) - L^r_s &\geq 0
\end{align*}
where $M\gg0$ (big-M) and $c \in \{0,1\}$ is responsible for making a choice between the two constraints.

\end{document}